\documentclass{article}

\usepackage[nonatbib,preprint]{neurips_2024}

\usepackage[utf8]{inputenc} 
\usepackage[T1]{fontenc}    
\usepackage{hyperref}       
\usepackage{url}            
\usepackage{booktabs}       
\usepackage{amsfonts}       
\usepackage{nicefrac}       
\usepackage{microtype}      
\usepackage{xcolor}         

\usepackage{algorithm}
\usepackage{algpseudocode}
\usepackage{amsmath}
\usepackage{graphicx}
\usepackage{multirow}
\usepackage{enumitem}
\usepackage{bm}

\title{SMART: Towards Pre-trained Missing-Aware Model for Patient Health Status Prediction}

\author{%
  Zhihao Yu$^1$ \quad Xu Chu$^{1,2,3}$ \quad Yujie Jin$^1$ \quad Yasha Wang$^{1,2}$ \quad Junfeng Zhao$^{1,2}$\\
  $^1$School of Computer Science, Peking University \\
  $^2$National Research and Engineering Center of Software Engineering, Peking University \\
  $^3$Center on Frontiers of Computing Studies, Peking University, Beijing, China\\
  \texttt{yuzhihao@stu.pku.edu.cn, chu\_xu@pku.edu.cn, wangyasha@pku.edu.cn} \\
}

\begin{document}

\maketitle

\begin{abstract}
  Electronic health record (EHR) data has emerged as a valuable resource for analyzing patient health status. However, the prevalence of missing data in EHR poses significant challenges to existing methods, leading to spurious correlations and suboptimal predictions. While various imputation techniques have been developed to address this issue, they often obsess unnecessary details and may introduce additional noise when making clinical predictions.
  To tackle this problem, we propose SMART, a Self-Supervised Missing-Aware RepresenTation Learning approach for patient health status prediction, which encodes missing information via elaborated attentions and learns to impute missing values through a novel self-supervised pre-training approach that reconstructs missing data representations in the latent space. By adopting missing-aware attentions and focusing on learning higher-order representations, SMART promotes better generalization and robustness to missing data. We validate the effectiveness of SMART through extensive experiments on six EHR tasks, demonstrating its superiority over state-of-the-art methods. 
\end{abstract}

\section{Introduction}

The rapid accumulation of electronic health record (EHR) data, driven by the widespread adoption of health information systems, has opened up new avenues for analyzing patient health status. EHR data in Intensive Care Units (ICUs) primarily captures patients' laboratory tests and vital signs, providing a rich resource for describing and analyzing their health conditions. These time series data have been leveraged to predict patient prognosis and physical status, attracting significant attention from both computer scientists and medical researchers. Various deep learning methods have been developed to exploit the patterns from EHR data \cite{song2018attend,luo2020hitanet,ma2021distilling,xu2023vecocare,xu2023seqcare,jiang2024graphcare,landi2020deep,miotto2016deep,gao2020stagenet}, aiming to assist doctors in making informed decisions, improving work efficiency, and ultimately enhancing patient outcomes.

Recent advancements in EHR analysis can be divided into two categories. The first category focuses on improving the classification ability of time series by capturing feature correlations and exploring the collaborative relationships between variables with various techniques, such as convolutions \cite{ma2020adacare}, attentions \cite{choi2016retain,baytas2017patient,ma2020concare,bai2018interpretable,ma2023mortality}, and graph neural networks \cite{zhang2021grasp,ma2022patient}. These methods have achieved promising results in patient health status prediction tasks, providing valuable insights into modeling the underlying patterns. However, a critical challenge is the prevalence of missing data in EHRs. Due to the fact that patients do not undergo all tests during each visit, EHR data is often highly sparse. These missing values can compromise the integrity of learned representations and potentially mislead models to capture spurious correlations, leading to erroneous predictions of patient health status. 
To address this issue, the other works \cite{shukla2021multi,rubanova2019latent,kim2023probabilistic} attempt to interpolate missing values in the input space by capturing the dynamics and regularizing the time series. Nevertheless, these methods may struggle with the complexity of the EHR data and concentrate on unnecessary details instead of capturing the implicit semantic information.
Furthermore, while these methods have shown promising results in imputing missing values and improving the performance of clinical tasks in their settings, they face challenges in fully utilizing the data for prediction since some observations are removed as supervised signals for imputation. Incorrect imputations may introduce additional noise and lead to unwanted distribution shifts, hindering the performance of subsequent tasks \cite{zhang2022graph}. 

To tackle these challenges, we propose SMART, a Self-Supervised Missing-Aware RepresenTation Learning approach for patient health status prediction. The primary objective of our approach is to enable the model to encode missing information effectively. It performs variable-independent encoding on multivariate time series inputs and utilizes Missing-Aware RepresenTation learning blocks, namely MART blocks, to capture temporal and variable interactions, with both modules incorporating missing information. Instead of deploying the vanilla self-attention mechanism \cite{vaswani2017attention}, we develop novel designs in attentions inside the MART block to cope with highly sparse data. Through stacking multiple blocks, the model can adequately learn correlations while perceiving missingness.

Another key innovation of our method lies in its two-stage training strategy. Inspired by previous work \cite{shukla2021multi,shukla2019interpolation,miao2021generative,chowdhury2023primenet} on imputing missing values, we adopt a self-supervised pre-training stage to enhance the ability to cope with sparse data, after which we fine-tune the model to accomplish the EHR tasks, such as mortality prediction. In contrast to previous works that perform reconstruction in the input space, our model conducts missing data reconstruction in the latent space. By randomly removing a portion of observations and training the model to reconstruct its representations, we enable the model to focus on learning higher-order representations which contain more semantic features rather than struggling with difficult-to-interpolate details. This approach promotes better generalization and robustness to missing data. After pre-training, the embedding decoder for reconstruction is replaced with a task-specific decoder for patient health status prediction during fine-tuning. To further bridge the gap between these two tasks, we introduce a learnable vector that serves as the query for the proposed attention mechanism and as the basis for patient health status prediction.

We validate the effectiveness of SMART through extensive experiments on six EHR tasks, including in-hospital mortality, sepsis, decompensation, phenotyping, and length of stay. Our results demonstrate that SMART, with its specific designs tailored to the missing characteristic of EHR data, significantly outperforms existing methods on all the metrics, setting a new state-of-the-art on these tasks. Furthermore, we showcase the robustness of our model under settings with higher missing rates, highlighting its potential for real-world applications in healthcare. Comparisons on the model efficiency illustrate that the proposed SMART is highly efficient, with lightweight parameters and fast training time among existing methods. Our work underscores the importance of integrating missing data awareness into representation learning for enhancing patient health status prediction and paves the way for more accurate and reliable clinical decision support systems.


\section{Related Work}
\textbf{Clinical Predictive Models for EHR:} Analyzing EHR data has become an increasingly popular research topic in the medical domain \cite{ma2023mortality,cai2016real}. Numerous deep learning models have been developed to mine and leverage information from EHR data \cite{ma2018health,si2021deep,zhang2022m3care,gao2019camp,bai2019improving}. Early methods used recurrent networks and attentions to capture temporal information \cite{choi2016retain,li2017blood,xu2018raim,ma2017dipole}. For learning better representation and achieving higher performance, some works attempt to learn feature correlations via sophisticated network architectures \cite{zang2021scehr,feng2021completing,lu2022context}. For example, Baytas et al. \cite{baytas2017patient} and Gao et al. \cite{gao2020stagenet} refine the design of recurrent networks by incorporating sampling intervals and disease progression in EHR data, thereby learning more comprehensive associations. Zhang et al. \cite{zhang2018patient2vec} design hierarchical recurrent encoders to capture both fragmented and global temporal variance. Ma et al. \cite{ma2020concare} and Ma et al. \cite{ma2023mortality} adopt attention mechanisms to enhance biomarkers that have strong connections with outcomes.
Other works try to incorporate medical knowledge from EHR data or human expertise \cite{jiang2024graphcare,choi2017gram,ma2018kame,gao2022medml}. For example, Zhang et al. \cite{zhang2021grasp} and Yu et al. \cite{yu2024predict} discover similar patients in the dataset and utilize their information to enhance learned representations and provide interpretations. Xu et al. \cite{xu2023vecocare} and Lu et al. \cite{lu2021collaborative} combine information from knowledge graphs for medical code data to improve predictions. Ye et al. \cite{ye2021medpath} integrate medical knowledge graphs with patient disease progression paths to obtain better health representations.
However, as discussed in the Introduction, all these methods face challenges in handling missing values. Most of them merely populate missingness with mean or front values during EHR data pre-processing, although it may be implausible and mislead the model to make wrong decisions. To address this challenge, we introduce missing-awareness mechanisms at multiple positions in SMART, enabling it to encode missingness and avoid the negative impact of missing values on representation learning.

\textbf{Imputation Models for EHR Analysis:} Imputation models are widely used in EHR analysis to handle missing values. 
Some works make efforts to interpolate missingness for learning better representations via clinical predictive tasks \cite{zhang2022graph,ma2020adversarial,tan2020data}. Specifically, Neil et al. \cite{neil2016phased}, Che et al. \cite{che2018recurrent}, and Tan et al. \cite{tan2020data} incorporate missing information into recurrent methods. Horn et al. \cite{horn2020set} use a set-based approach and transform time series into sets of observations modeled by set functions insensitive to misalignment. Zhang et al. \cite{zhang2023warpformer} consider sampling frequency and unify irregular time series in multiple scales. Nevertheless, these methods do not apply reconstruction losses and do not perform real imputations.
The other works impute irregularly observed values to aligned reference points, notably recurrent methods \cite{miao2021generative}, variational auto-encoders \cite{shukla2019interpolation,shukla2021multi}, generative adversarial networks \cite{miao2021generative,zhang2021missing}, ordinary differential equations \cite{rubanova2019latent,chen2024contiformer}, and probabilistic interpolation methods \cite{kim2023probabilistic}.
However, this paradigm randomly masks existing observations and conducts both clinical tasks and interpolation simultaneously, making already sparse data even sparser when predicting health status, leading to unsatisfactory performance compared to methods that do not mask observations.
Recently, a pre-training approach \cite{chowdhury2023primenet} has been proposed for empowering models to impute missing values through self-supervised pre-training and then perform clinical tasks, allowing the model to execute EHR analyses on intact data while possessing imputation capabilities. Nonetheless, interpolating in the input space can lead to getting bogged down in optimizing some details rather than capturing the implicit information in the entire sequence, as well as skewing the underlying data distribution \cite{zhang2022graph}. In contrast, SMART conducts missing data reconstruction in the latent space, enabling the model to focus on learning more semantic representations.

\section{Methodology}
\begin{figure}[h]
    \centering
    \includegraphics[width=0.9\columnwidth]{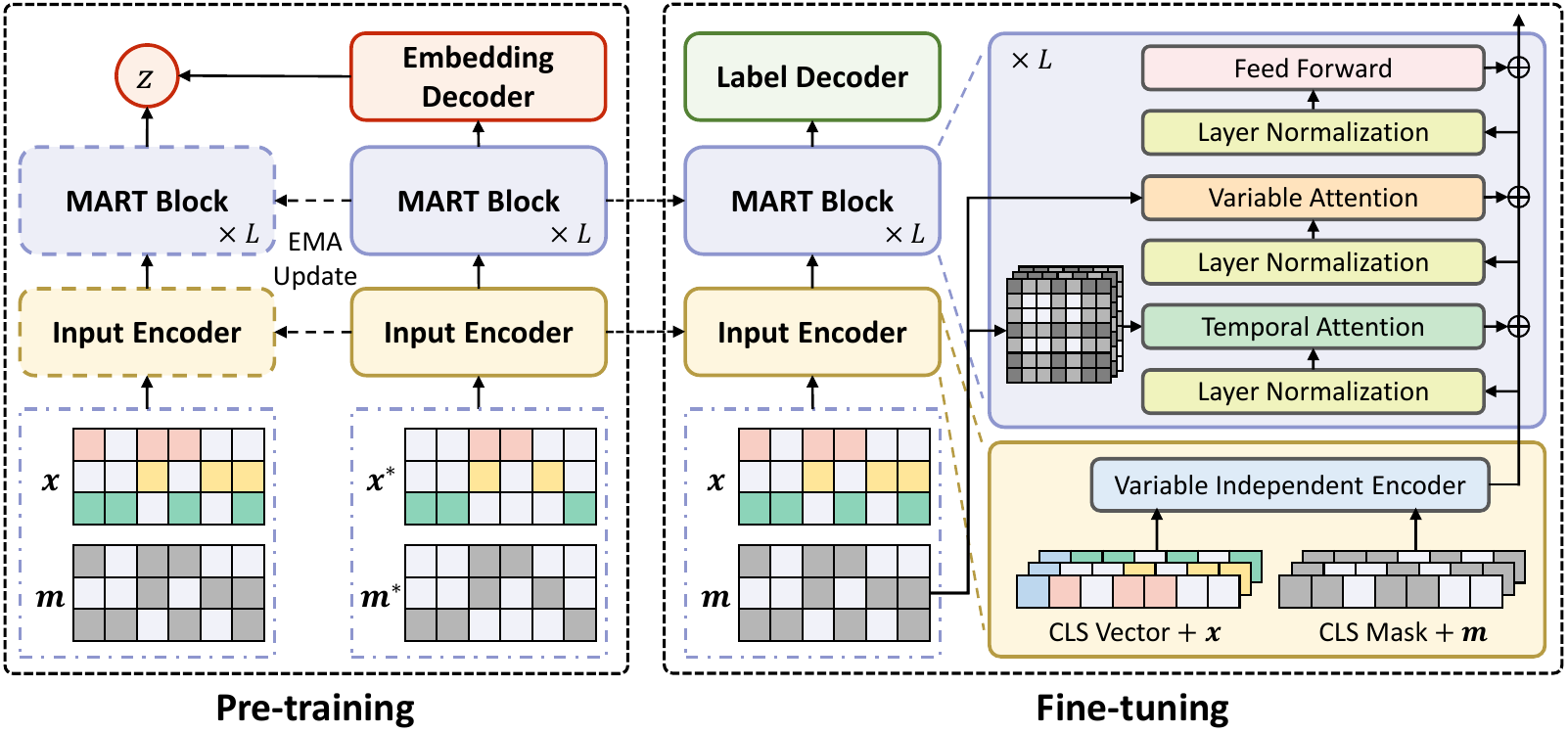}
    \caption{\textbf{Overview of SMART}. \textbf{\textit{Left}}: We randomly mask EHR data and conduct reconstruction in the latent space. The reconstruction targets are generated by EMA updated parameters. \textbf{\textit{Right}}: We illustrate the detailed architecture of the input encoder and the MART block. The input encoder embeds each variable (which can also be referred to as a biomarker) and missing mask into a separate hidden space. The MART block employs various techniques to capture feature interactions in both the temporal and variable dimensions while further encoding missing information.}
    \label{framework}
\end{figure}

In this section, we present the proposed SMART, a self-supervised missing-aware representation learning approach for predicting patients' health status. In Figure \ref{framework}, we give an overview of the information flow and highlight our designs.

\subsection{Variable Independent Encoder}
\textbf{Notation:} Let $\displaystyle (\bm{x}, \bm{m}, y)$ denotes the EHR data of a specific patient, where the visit sequence $\displaystyle \bm{x} \in \mathbb{R}^{T \times N}$ contains $T$ visits with $N$ variables. $\displaystyle \bm{x}^{t}_{n}$ records the value of the $n$-th indicator for the patient at visit $t$, accompanied by a binary mask $\displaystyle \bm{m}^{t}_{n}$ indicating whether the value is observed. For different patients, the visit length $T$ can vary. To avoid the possible negative effects of varying intervals between visits, the observation intervals are aligned to hours following \cite{ma2020concare,zhang2021grasp,ma2022patient,yu2024predict}. Every patient corresponds to a label $y$, indicating the diagnosis or the outcome. Our goal is to improve the model's representation learning capability to achieve better classification performance on $y$.

Given a patient's EHR data, we apply a variable-independent encoding strategy to map $(\bm{x}, \bm{m})$ into a latent representation $\displaystyle \bm{h} \in \mathbb{R}^{T \times N \times d}$, inspired by the success of previous works \cite{ma2023mortality,zhang2023warpformer,nie2023time}, where $d$ is the dimension of latent space. Although variable-dependent approaches \cite{gao2020stagenet,ma2020adacare} capture variable interactions and compress dimensions when embedding, such methods cannot learn further associations, i.e., the attention weight of each variable \cite{ma2020concare}. Unlike variable-independent recurrent models \cite{ma2020concare,ma2022patient,yu2024predict}, we adopt linear projections to encode data, allowing the entire encoding process to be parallel and without accumulating noise from past visits. When handling variable values $\bm{x}$ and masks $\bm{m}$, we combine them directly and let the model capture the interactions between them.

\textbf{CLS Vector:} When encoding the EHR data, we introduce a learnable vector which is concatenated before the time series. The role of this vector is to learn the information of the whole sequence in subsequent interactions and act as the pooled hidden state used for classification (similar to the $\texttt{[CLS]}$ token in the language model such as BERT \cite{kenton2019bert}). This design also bridges the gap between pre-training and fine-tuning tasks, as the information in the vector is used for both reconstruction and prediction. Specifically, we concatenate the CLS vector $\bm{v} \in \mathbb{R}^{N \times d}$ at the temporal dimension before the hidden representation $\bm{h}$, which can be formulated as $\displaystyle \bm{h}_v = [\bm{v}, \bm{h}]$, where $\displaystyle \bm{h}_v \in \mathbb{R}^{(T+1) \times N \times d}$. This process can be interpreted as concatenating a learnable parameter before the time series $\bm{x}$ and a $\texttt{True}$ vector before the mask $\bm{m}$, making the consultation sequence one step longer (as illustrated in Figure \ref{framework}). Here, we concatenate on the embedding $\bm{h}$ instead of adding vectors to the input data since this procedure provides more flexibility in the learned representation space.
At the end of this module, we introduce positional information to the representation $\bm{h}_v$ for subsequent interactions in the MART blocks using sinusoidal positional encoding \cite{vaswani2017attention}.

\subsection{MART Block}
The MART block is the core module elaborated to learn the patient's health representation. It is mainly composed of two attention mechanisms operating on the temporal and variable dimensions. To further mitigate the impact of missing data on representation learning, we introduce masks into the attention mechanism and strengthen the attention weights of existing observations.

\textbf{Temporal Attention:} The variable-independent encoder described above embeds patient information into representation $\displaystyle \bm{h}_v \in \mathbb{R}^{(T+1) \times N \times d}$. In this section, we introduce how we calculate temporal weights leveraging mask information. Following the convention for self-attention mechanisms, we compute the query, key, and value via linear transformations of $\displaystyle \bm{h}_v$ for the temporal attention, denoted as $\bm{Q}_{temp}$, $\bm{K}_{temp}$, and $\bm{V}_{temp}$. To incorporate mask information, we construct the temporal attention bias $\bm{B} \in \mathbb{R}^{(T+1) \times (T+1) \times N}$. The bias $\bm{B}_n^{i,j}$ between visit $i$ and $j$ for the $n$-th variable can be computed by
\begin{equation}
    \bm{m}_n' = [\texttt{True}, \bm{m}_n],\\
\end{equation}
\begin{equation}
    \bm{B}_n^{i,j} = \begin{cases}
        2,\; &{\bm{m}'}_n^i = \texttt{True} \; \texttt{and} \; {\bm{m}'}_n^j = \texttt{True},\\
        1\; &{\bm{m}'}_n^i \; \texttt{xor} \; {\bm{m}'}_n^j = \texttt{True},\\
        0\; &{\bm{m}'}_n^i = \texttt{False} \; \texttt{and} \; {\bm{m}'}_n^j = \texttt{False},\\
    \end{cases}
\end{equation}
where $[\cdot, \cdot]$ denotes concatenation of matrices. Then we obtain the temporal attention weights through $\displaystyle \texttt{softmax}\Big(\frac{\bm{Q}_{temp}\bm{K}_{temp}^\top}{\sqrt{d}} + \bm{B}\Big)$, where both matrix multiplication and softmax are imposed on the temporal dimension.
In this way, we strengthen information from observed visits and suppress the others. At the same time, we do not completely block the missing visits, providing an opportunity for the model to interpolate them and utilize their information.

\textbf{Variable Attention:} We capture correlations between variables via the proposed variable attention. In contrast to previous approaches which calculate variable interactions separately for each visit  \cite{zhang2023warpformer,nie2023time}, we capture variable relationships from a global perspective of the patient. Given representation $\bm{h}_{temp} \in \mathbb{R}^{(T+1) \times N \times d}$ from the temporal attention, we get query, key, and value $\bm{Q}_{var}, \bm{K}_{var}, \bm{V}_{var}$ for the variable attention mechanism as follows:
\begin{equation}
\begin{split}
    \bm{Q}_{var} &= \texttt{Linear}(\bm{h}_{temp}^0),\\
    \bm{K}_{var} &= \texttt{Linear}\Big(\sum_{t}\bm{h}_{temp}^t \;\; \texttt{where} \;{\bm{m}'}^t=\texttt{True}\Big),\\
    \bm{V}_{var} &= \texttt{Linear}(\bm{h}_{temp}).\\
\end{split}
\end{equation}

The query $\bm{Q}_{var}$ is only obtained using the vector at the first step ($\bm{h}_{temp}^0$), which is also the location of learnable parameter $\bm{v}$ inserted in the input encoder. This operation motivates these vectors to learn the overall health state of the patient. The key $\bm{K}_{var}$ are acquired using the averaged embedding of all observed visits to minimize the effect of missingness in visits. Then we calculate the time-invariant correlations between the variables and get attention output by $\displaystyle \texttt{softmax}\Big(\frac{\bm{Q}_{var}\bm{K}_{var}^\top}{\sqrt{d}}\Big)\bm{V}_{var}$.

Between these attentions, we utilize layer normalizations \cite{ba2016layer} and skip connections \cite{he2016deep} to avoid overfitting and accelerate convergence. Besides, a feed-forward constituted by linear projections and activation functions is used following the vanilla design of Transformer \cite{vaswani2017attention}. The MART block can be stacked in multiple layers to allow for sufficient interactions. The final health status of patients embedded by the MART blocks is denoted as $\bm{s} \in \mathbb{R}^{(T+1) \times N \times d}$.

\subsection{Two-Stage Training Strategy}
\textbf{Pre-training Stage:} To empower the model with missing imputation capabilities to enhance the learned representation, we propose a self-supervised pre-training method that occludes some of the observations and reconstructs their representation in the hidden space. Different from previous methods \cite{chowdhury2023primenet}, we do not seek complex manual data augmentation, but simply remove some of the observations as the targets whose representations will be reconstructed. When generating the targets, we generate them randomly with a probability interval rather than with a fixed probability. This encourages the model to achieve better generalization across sequences with different sampling rates, rather than overfitting on missing probabilities. For the purpose of strengthening the reconstructing capability additionally and avoiding the model being trapped in the local minimum, we randomly sample the targets in each epoch of pre-training instead of using fixed data.

To emphasize a nontrivial pre-training task, we apply a self-motivated paradigm with an asymmetric architecture inspired by \cite{assran2023self}, as illustrated in Figure \ref{framework}. Specifically, given the EHR data $(\bm{x}, \bm{m})$, we randomly generate a mask $\bm{\hat{m}}$ to remove the existing observations partially and obtain augmented data $(\bm{x}^*, \bm{m}^*)$. Defining the modules being trained (input encoder, MART blocks, and embedding decoder) as $f$ and the modules (input encoder and MART blocks) used to generate labels as $\hat{f}$, the reconstructions $\bm{s}^*_{pre\textrm{-}train}$ are generated by $f$ using augmented data, while the reconstruction targets $\hat{\bm{s}}$ are acquired by embedding the original data via $\hat{f}$ whose parameters are updated by exponential moving average (EMA) \cite{hunter1986exponentially} of the parameters from $f$. The embedding decoder consists of an MLP with activation functions that uses the patient health representation $\bm{s}^* \in \mathbb{R}^{(T+1) \times N \times d}$ from augmented data as input and outputs the reconstructed health state $\bm{s}^*_{pre\textrm{-}train} \in \mathbb{R}^{(T+1) \times N \times d}$. This paradigm provides a smooth label update curve that avoids model oscillations and underfitted embedding decoder. The pre-training loss is computed by $L_1$ distance considering only the features of the removed data (i.e., the position where $\hat{\bm{m}} = \texttt{True}$) in the latent space:
\begin{equation}
    \mathcal{L}_{pre\textrm{-}train} = \sum \hat{\bm{m}} \cdot \|\bm{s}^*_{pre\textrm{-}train} - \hat{\bm{s}}\|_1.
\end{equation}
Since the model is trained to reconstruct the missing data, it is forced to learn the underlying structure of the data and the temporal relationships between variables, which can be beneficial for subsequent tasks. In particular, the CLS vector is not aligned in $\mathcal{L}_{pre\textrm{-}train}$ (i.e., the position of the CLS vector is $\texttt{False}$ in the mask $\bm{\hat{m}}$), which allows the model to store the information of the whole sequence in the CLS vector and exploit it in the fine-tuning stage.

\textbf{Fine-tuning Stage:} After pre-training, we replace the embedding decoder with a task-specific decoder for patient health status prediction, such as classification task. While making predictions, we only use the representation at the first step ($\bm{s}^0$), which is the position of the CLS vector. The label decoder is an MLP with layer normalizations and activations which can be simplified as a projection function $\mathbb{R}^{N \times d} \rightarrow \mathbb{R}^{|y|}$. In the fine-tuning stage, the parameters are updated by the task-specific loss (e.g. cross-entropy for classification).
In the first few rounds of training, we freeze the parameters of the other modules and only update the label decoder to make the pre-trained parameters to be reserved. The proposed two-stage training procedure can be formulated in Algorithm \ref{alg:SMART}:

\begin{algorithm}
\caption{Algorithm of SMART}
\label{alg:SMART}
\begin{algorithmic}[1]
    \ForAll{$E$ in pre-training epochs}
        \State Sample a mask $\hat{\bm{m}}$ with a probability interval $p$ and generate augmented data $(\bm{x}^*, \bm{m}^*)$
        \State Generate reconstructions $\displaystyle \bm{s}^*_{pre\textrm{-}train} \leftarrow f(\bm{x}^*, \bm{m}^*)$ and reconstruction targets $\bm{\hat{s}}\leftarrow \hat{f}(\bm{x}, \bm{m})$
        \State Update $f$ by minimizing $\displaystyle \mathcal{L}_{pre\textrm{-}train}$
        \State Update $\hat{f} \leftarrow \texttt{EMA}(\hat{f}, f)$
    \EndFor
    \State Freeze the parameters of the input encoder and the MART blocks
    \ForAll{$E$ in fine-tuning epochs}
        \If{$E$ = unfreeze epoch}
            \State Unfreeze the parameters of the input encoder and the MART blocks
        \EndIf
        \State Update parameters by minimizing the task-specific loss with prediction
    \EndFor
\end{algorithmic}
\end{algorithm}



\section{Experiments}
  
\subsection{Experimental Settings}
\textbf{Datasets:} We follow previous works \cite{ma2020adacare,ma2020concare,ma2022patient,zhang2023warpformer} to compare models on three EHR datasets, Cardiology \cite{goldberger2000physiobank,silva2012predicting}, Sepsis \cite{reyna2020early}, and MIMIC-III \cite{johnson2016mimic}. The \textbf{Cardiology} dataset consists of 37 vital signs and biomarkers describing patients admitted to cardiac, medical, surgical, and trauma ICUs. Each record contains sparse measurements from the first 48 hours after admission. We follow the preprocessing procedures of previous works \cite{ma2020concare,zhang2022graph} where the observation times are aligned to hours. After preprocessing, there are 11,988 patients and the observed rate is 24.7\%. The prediction target is to determine in-hospital mortality. 13.8\% of examples are in the positive class.
The \textbf{Sepsis} dataset contains 34 vital signs and laboratory values relevant to sepsis. The EHR data are recorded once an hour including 40,335 patients from ICUs. The prediction task is to identify the patient's risk of developing sepsis, with a positive rate of 7.3\%. For convenience, we used only the records of the first 60 visits for each patient. The observed rate is only 19.8\%.
The \textbf{MIMIC-III} dataset is a multivariate time series dataset consisting of 17 physiological signals after pre-processing. It contains many clinical scenarios calling for accurate predictions over diversified clinical signals. We conduct four clinical tasks on the MIMIC-III including \textbf{in-hospital mortality}, \textbf{decompensation}, \textbf{phenotyping}, and \textbf{length of stay}. Additional details on these datasets can be found in Appendix \ref{sec:dataset}.

\begin{table}
    \caption{Performance comparison with standard deviation on the six clinical tasks. The best results are marked in bold. The second best results are underlined.}
    \label{result}
    \centering
    \small
    \resizebox{\linewidth}{!}{
    \begin{tabular}{l|cccccc}
        \toprule
        \multirow{2}{*}{\textbf{Model}} & \multicolumn{2}{c}{\textbf{Cardiology}} & \multicolumn{2}{c}{\textbf{Sepsis}} & \multicolumn{2}{c}{\textbf{In-hospital Mortality}}\\
        & AUPRC(\%) & F1 Score(\%) & AUPRC(\%) & F1 Score(\%) & AUPRC(\%) & F1 Score(\%)\\
        \midrule
        AdaCare & 40.54{\scriptsize $\pm$2.36} & 27.80{\scriptsize $\pm$2.97} & 41.13{\scriptsize $\pm$2.75} & 36.87{\scriptsize $\pm$0.96} & 41.26{\scriptsize $\pm$1.62} & 27.47{\scriptsize $\pm$1.96} \\
        StageNet & 47.01{\scriptsize $\pm$2.94} & 38.27{\scriptsize $\pm$2.74} & 59.11{\scriptsize $\pm$4.63} & 54.89{\scriptsize $\pm$3.28} & 51.77{\scriptsize $\pm$3.62} & \underline{40.90{\scriptsize $\pm$1.69}} \\
        ConCare & 47.77{\scriptsize $\pm$3.62} & 30.22{\scriptsize $\pm$2.79} & \underline{74.83{\scriptsize $\pm$3.47}} & 67.27{\scriptsize $\pm$3.46} & 51.64{\scriptsize $\pm$3.34} & 35.76{\scriptsize $\pm$2.58} \\
        GRASP & 47.93{\scriptsize $\pm$5.73} & 31.90{\scriptsize $\pm$2.60} & 74.47{\scriptsize $\pm$3.79} & 66.75{\scriptsize $\pm$3.41} & \underline{51.84{\scriptsize $\pm$3.42}} & 38.00{\scriptsize $\pm$1.75} \\
        SAFARI & 45.58{\scriptsize $\pm$3.59} & 11.06{\scriptsize $\pm$2.09} & 72.06{\scriptsize $\pm$1.74} & \underline{69.06{\scriptsize $\pm$1.14}} & 48.96{\scriptsize $\pm$3.01} & 33.96{\scriptsize $\pm$3.83} \\
        PPN & \underline{49.98{\scriptsize $\pm$3.97}} & 37.64{\scriptsize $\pm$3.19} & 72.70{\scriptsize $\pm$4.99} & 68.08{\scriptsize $\pm$2.71} & 51.24{\scriptsize $\pm$2.96} & 40.60{\scriptsize $\pm$1.83} \\
        RainDrop & 37.23{\scriptsize $\pm$1.20} & 23.85{\scriptsize $\pm$3.39} & 72.82{\scriptsize $\pm$3.68} & 67.58{\scriptsize $\pm$2.54} & 46.21{\scriptsize $\pm$2.71} & 36.29{\scriptsize $\pm$3.20} \\
        Warpformer & 48.10{\scriptsize $\pm$3.95} & \underline{40.26{\scriptsize $\pm$3.68}} & 73.03{\scriptsize $\pm$2.53} & 65.93{\scriptsize $\pm$1.79} & 51.49{\scriptsize $\pm$1.89} & 36.83{\scriptsize $\pm$3.12} \\
        PrimeNet & 49.25{\scriptsize $\pm$2.36} & 37.10{\scriptsize $\pm$2.67} & 33.38{\scriptsize $\pm$2.29} & 12.46{\scriptsize $\pm$2.87} & 51.41{\scriptsize $\pm$0.89} & 38.71{\scriptsize $\pm$2.62} \\
        \midrule
        SMART & \textbf{53.84{\scriptsize $\pm$2.24}} & \textbf{47.53{\scriptsize $\pm$2.33}} & \textbf{81.67{\scriptsize $\pm$0.84}} & \textbf{75.37{\scriptsize $\pm$2.62}} & \textbf{53.30{\scriptsize $\pm$0.12}} & \textbf{44.23{\scriptsize $\pm$2.03}} \\
        \midrule
        \multirow{2}{*}{\textbf{Model}} & \multicolumn{2}{c}{\textbf{Decompensation}} & \multicolumn{2}{c}{\textbf{Phenotyping}} & \multicolumn{2}{c}{\textbf{Length of Stay}}\\
        & AUPRC(\%) & F1 Score(\%) & ma-ROC(\%) & mi-ROC(\%) & ma-ROC(\%) & mi-ROC(\%)\\
        \midrule
        AdaCare & 52.41{\scriptsize $\pm$1.76} & 51.62{\scriptsize $\pm$0.84} & 63.28{\scriptsize $\pm$0.62} & 73.75{\scriptsize $\pm$0.26} & 66.06{\scriptsize $\pm$0.97} & 82.11{\scriptsize $\pm$0.40} \\
        StageNet & 64.97{\scriptsize $\pm$0.60} & 62.51{\scriptsize $\pm$0.99} & 73.33{\scriptsize $\pm$0.30} & 79.53{\scriptsize $\pm$0.31} & \underline{69.54{\scriptsize $\pm$0.39}} & 83.26{\scriptsize $\pm$0.15} \\
        ConCare & 65.50{\scriptsize $\pm$1.30} & 59.73{\scriptsize $\pm$2.25} & 71.05{\scriptsize $\pm$0.36} & 78.03{\scriptsize $\pm$0.21} & 69.27{\scriptsize $\pm$0.50} & 83.19{\scriptsize $\pm$0.12} \\
        GRASP & 64.29{\scriptsize $\pm$0.64} & 58.48{\scriptsize $\pm$0.13} & 69.08{\scriptsize $\pm$0.47} & 76.97{\scriptsize $\pm$0.25} & 69.24{\scriptsize $\pm$0.47} & 83.20{\scriptsize $\pm$0.20} \\
        SAFARI & 61.93{\scriptsize $\pm$1.57} & 59.71{\scriptsize $\pm$0.11} & 66.91{\scriptsize $\pm$0.84} & 75.69{\scriptsize $\pm$0.45} & 68.51{\scriptsize $\pm$0.48} & 82.93{\scriptsize $\pm$0.21} \\
        PPN & 64.91{\scriptsize $\pm$0.33} & 62.80{\scriptsize $\pm$1.41} & 68.77{\scriptsize $\pm$0.54} & 76.55{\scriptsize $\pm$0.41} & 68.52{\scriptsize $\pm$0.48} & 83.13{\scriptsize $\pm$0.18} \\
        RainDrop & 61.16{\scriptsize $\pm$1.47} & 58.21{\scriptsize $\pm$2.61} & 71.59{\scriptsize $\pm$0.44} & 78.38{\scriptsize $\pm$0.42} & 68.23{\scriptsize $\pm$0.66} & 82.83{\scriptsize $\pm$0.18} \\
        Warpformer & \underline{68.21{\scriptsize $\pm$1.64}} & \underline{63.70{\scriptsize $\pm$0.88}} & \underline{74.65{\scriptsize $\pm$0.23}} & \underline{80.36{\scriptsize $\pm$0.23}} & 69.53{\scriptsize $\pm$0.45} & \underline{83.34{\scriptsize $\pm$0.28}} \\
        PrimeNet & 57.86{\scriptsize $\pm$1.32} & 51.59{\scriptsize $\pm$1.71} & 68.90{\scriptsize $\pm$0.17} & 76.66{\scriptsize $\pm$0.19} & 67.88{\scriptsize $\pm$0.40} & 82.89{\scriptsize $\pm$0.18} \\
        \midrule
        SMART & \textbf{71.26{\scriptsize $\pm$0.81}} & \textbf{67.38{\scriptsize $\pm$0.67}} & \textbf{76.13{\scriptsize $\pm$0.14}} & \textbf{81.41{\scriptsize $\pm$0.18}} & \textbf{70.29{\scriptsize $\pm$0.44}} & \textbf{83.92{\scriptsize $\pm$0.24}} \\
        \bottomrule
    \end{tabular}
    }
\end{table}

\textbf{Evaluation Protocols:} We assess the performance on the binary classification tasks (including Cardiology, Sepsis, in-hospital mortality, and decompensation) using the area under the precision-recall curve (AUPRC) and F1 Score. AUPRC is the most informative and primary evaluation metric when dealing with a highly imbalanced and skewed dataset \cite{davis2006relationship} such as healthcare data. We calculate F1 Score focusing on positive patients, which is more relevant in clinical scenarios. Phenotyping is a multi-label classification, each label indicating the diagnosis of a phenotype. Therefore, we examine phenotyping on the macro and micro area under the receiver operating characteristic curve (AUROC), abbreviated as ma-ROC and mi-ROC, respectively. For length-of-stay prediction, we follow Harutyunyan et al. \cite{Harutyunyan2019} to separate labels into multiple bins and evaluate the ma-ROC and mi-ROC.
We randomly divide the data set into a training set containing 80\% of the patients, a validation set of 10\% patients, and a test set containing the remaining 10\% instances. The model achieving the best AUPRC (or ma-ROC) on the validation set is evaluated on the test set. To eliminate the randomness, we conduct each experiment with three random seeds and report both the mean and standard deviation of the results. Results on more metrics are provided in Table \ref{more_results}. Hyperparameters and implementation details are introduced in Appendix \ref{sec:implementation}.

\textbf{Baselines:} We organize existing methods of handling EHR data into three paradigms. The first type utilizes various techniques to enhance representation learning of EHR data. Such as \textbf{AdaCare} \cite{ma2020adacare}, a multi-scale model focusing on extracting information from multiple levels, \textbf{StageNet} \cite{gao2020stagenet}, which refines the design of LSTM by incorporating personalized disease stage development, and \textbf{ConCare} \cite{ma2020concare}, which consists variable-independent encoders and self-attention mechanism to learn feature correlations. Works that employ global information or other patient information to aid in modeling also belong to this category, including \textbf{SAFARI}, which learns feature correlations from a group-wise perspective and integrates correlations by a graph neural network (GNN), \textbf{PPN}, a model that extracts typical patients and uses them help predicting and interpreting, and \textbf{GRASP}, which exploits similar patients and GNN to improve learned representation. The second paradigm is the state-of-the-art that delves into interpolating missingness and achieving better performance on clinical tasks, such as \textbf{RainDrop} \cite{zhang2022graph}, which utilizes GNN to capture the dependency among variables, \textbf{Warpformer} \cite{zhang2023warpformer}, which estimates the sampling frequency and interpolates missingness with attentions. The others are self-supervised pre-training methods, including \textbf{PrimeNet} \cite{chowdhury2023primenet}, a model that is aware of missingness and adopts input reconstruction with contrastive learning to pre-train the model.

\subsection{Experimental Results}
\subsubsection{Main Results}
We include the overall comparison results on the six tasks in Table \ref{result}. SMART consistently outperforms existing baselines across all prediction metrics. Specifically, on the binary classification tasks, SMART achieves the results with an average improvement of 3.80\% and 5.15\% absolutely on AUPRC and F1 Score compared to the best baseline, respectively. The results demonstrate the effectiveness of SMART in learning representations and predicting patient health status. AdaCare performs the worst in the baseline, which may be due to the fact that its convolutional structure only averages the proximity visits and does not perceive the missing ones. Interestingly, although StageNet and PrimeNet show competitive performance on some tasks such as Cardiology and in-hospital mortality, they perform surprisingly poorly on the Sepsis dataset. This is because the Sepsis dataset has a lower observed rate, which makes it more challenging for models to learn effective representations. Some recurrent models, including ConCare, GRASP, and PPN, exhibit robust performance across all datasets among the baselines. Nevertheless, with the capability of encoding missingness, SMART achieves remarkable performance improvements.
As we observed, on the datasets with higher missing rate (Cardiology and Sepsis), SMART outperforms the other methods by a larger margin, indicating its robustness to missing values. On the phenotyping and length-of-stay prediction, our proposed model achieves the best results as well, showing its generalization ability to different clinical scenarios. The results also show that the performance of SMART is more stable than other methods, as indicated by the smaller standard deviation.

\subsubsection{Ablation Study}
To evaluate the effectiveness of each component in SMART, we conduct an ablation study on the Cardiology, Sepsis, and in-hospital mortality. Firstly, we investigate the effects of different pre-training strategies on model performance. We introduce two variants of the self-supervised pre-training strategy: (1) imputing the missing values in the input space like previous methods \cite{shukla2021multi,miao2021generative,chowdhury2023primenet} (\textbf{w/ Imputation}), and (2) directly training the model for patient health status prediction without the proposed self-supervised pre-training (\textbf{w/o Pre-training}). As the results shown in Table \ref{ablation}, we observe that although combining the pre-training strategy in the input space can improve the performance compared to models without pre-training, it is still inferior to the proposed strategy that reconstructs in the representation space.  These findings highlight the necessity of the pre-training strategy that integrates the missing imputation ability by reconstructing latent representations.

Additionally, we explore the importance of different components and designs in SMART, and the results are also presented in Table \ref{ablation}. In particular, we compare with the following reduced variants: (1) \textbf{w/o Mask}, which removes the mask information totally in SMART, including mask in the input encoder and the attentions in the MART blocks; (2) \textbf{w/o Temporal Attention}, which removes the temporal attention mechanism in SMART; (3) \textbf{w/o Variable Attention}, which removes the variable attention mechanism in SMART; and (4) \textbf{w/o CLS Vector}, which removes the CLS vector in the input encoder and using the representation at last observation as query in the variable attention and prediction. We observe that all components significantly contribute to the improvement. Notably, incorporating missing information is important for the model to learn high-quality representations. The results show that the temporal and variable attention mechanisms are crucial and fundamental in capturing temporal and variable dependencies, respectively. The CLS vector also plays a critical role in improving the effect of pre-training since it narrows the difference between the two training stages. By considering these components together, SMART is able to capture the intricate temporal relationships and characteristics inner sparse EHR data.

\begin{table}
    \caption{Ablation study of SMART on the Cardiology, Sepsis, and MIMIC-III in-hospital mortality.}
    \label{ablation}
    \centering
    \small
    \resizebox{\linewidth}{!}{
    \begin{tabular}{l|cccccc}
        \toprule
        \multirow{2}{*}{\textbf{SMART}} & \multicolumn{2}{c}{\textbf{Cardiology}} & \multicolumn{2}{c}{\textbf{Sepsis}} & \multicolumn{2}{c}{\textbf{In-hospital Mortality}}\\
        & AUPRC(\%) & F1 Score(\%) & AUPRC(\%) & F1 Score(\%) & AUPRC(\%) & F1 Score(\%)\\
        \midrule
        Full & \textbf{53.84{\scriptsize $\pm$2.24}} & \textbf{47.53{\scriptsize $\pm$2.33}} & \textbf{81.67{\scriptsize $\pm$0.84}} & \textbf{75.37{\scriptsize $\pm$2.62}} & \textbf{53.30{\scriptsize $\pm$0.12}} & \textbf{44.23{\scriptsize $\pm$2.03}} \\
        \midrule
        w/ Imputation & 52.77{\scriptsize $\pm$2.07} & 42.40{\scriptsize $\pm$2.69} & 81.04{\scriptsize $\pm$2.84} & 74.71{\scriptsize $\pm$2.61} & 52.91{\scriptsize $\pm$1.42} & 43.69{\scriptsize $\pm$3.24} \\
        w/o Pre-training & 52.26{\scriptsize $\pm$3.14} & 46.60{\scriptsize $\pm$2.07} & 79.54{\scriptsize $\pm$3.22} & 74.26{\scriptsize $\pm$2.65} & 51.84{\scriptsize $\pm$0.95} & 41.69{\scriptsize $\pm$2.95} \\
        \midrule
        w/o Mask & 49.03{\scriptsize $\pm$1.93} & 41.98{\scriptsize $\pm$1.71} & 76.35{\scriptsize $\pm$2.76} & 69.51{\scriptsize $\pm$2.11} & 50.43{\scriptsize $\pm$2.44} & 39.70{\scriptsize $\pm$1.60} \\
        w/o Temporal Attention & 49.28{\scriptsize $\pm$2.94} & 40.53{\scriptsize $\pm$2.31} & 65.22{\scriptsize $\pm$1.44} & 58.47{\scriptsize $\pm$2.36} & 46.64{\scriptsize $\pm$1.63} & 33.37{\scriptsize $\pm$3.82} \\
        w/o Variable Attention & 52.27{\scriptsize $\pm$2.53} & 44.75{\scriptsize $\pm$2.82} & 80.47{\scriptsize $\pm$2.36} & 74.65{\scriptsize $\pm$2.39} & 50.98{\scriptsize $\pm$0.62} & 41.70{\scriptsize $\pm$2.92} \\
        w/o CLS Vector & 52.96{\scriptsize $\pm$0.34} & 46.42{\scriptsize $\pm$1.77} & 77.56{\scriptsize $\pm$2.89} & 71.53{\scriptsize $\pm$2.56} & 50.78{\scriptsize $\pm$1.64} & 43.86{\scriptsize $\pm$2.85} \\
        \bottomrule
    \end{tabular}
    }
\end{table}

\subsubsection{Effect of Missingness}
\begin{figure}
    \setlength{\abovecaptionskip}{-0.2cm}
    \setlength{\belowdisplayskip}{0cm}
    \centering
    \includegraphics[width=1.0\linewidth]{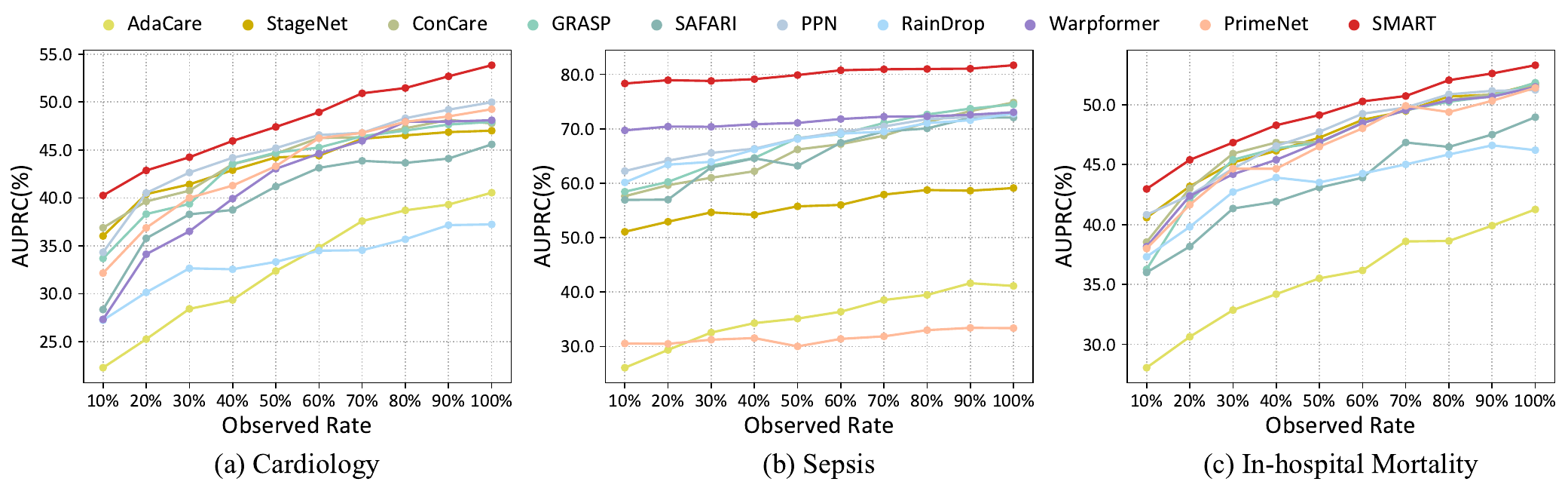}
    \caption{Performance on different observed ratio of EHR.}
    \label{missing_rate}
\end{figure}

To further investigate the impact of missingness in the data on performance, we conduct a comprehensive experiment on the Cardiology, Sepsis, and in-hospital mortality tasks by varying the observed rate from 10\% to 100\%. The results of AUPRC are shown in Figure \ref{missing_rate}. AdaCare shows the most significant performance degradation as the observed rate decreases, which confirms that the convolutional architecture inside AdaCare is sensitive to missing values. Though Warpformer is stable on the Sepsis, it cannot handle the missingness well on the Cardiology and in-hospital mortality. Among the baselines, StageNet and PPN exhibit robustness to missing values. However, there is still a large margin compared to SMART. Especially, on the Sepsis, we observe only a very small decline on SMART despite only 10\% of the available data.
SMART performs outstandingly under different missing rate scenarios, demonstrating its fruitful results in missingness perception. 

\subsection{Model Efficiency}
\begin{figure}
    \centering
    \setlength{\abovecaptionskip}{-0.2cm}
    \setlength{\belowdisplayskip}{0cm}
    \includegraphics[width=1.0\linewidth]{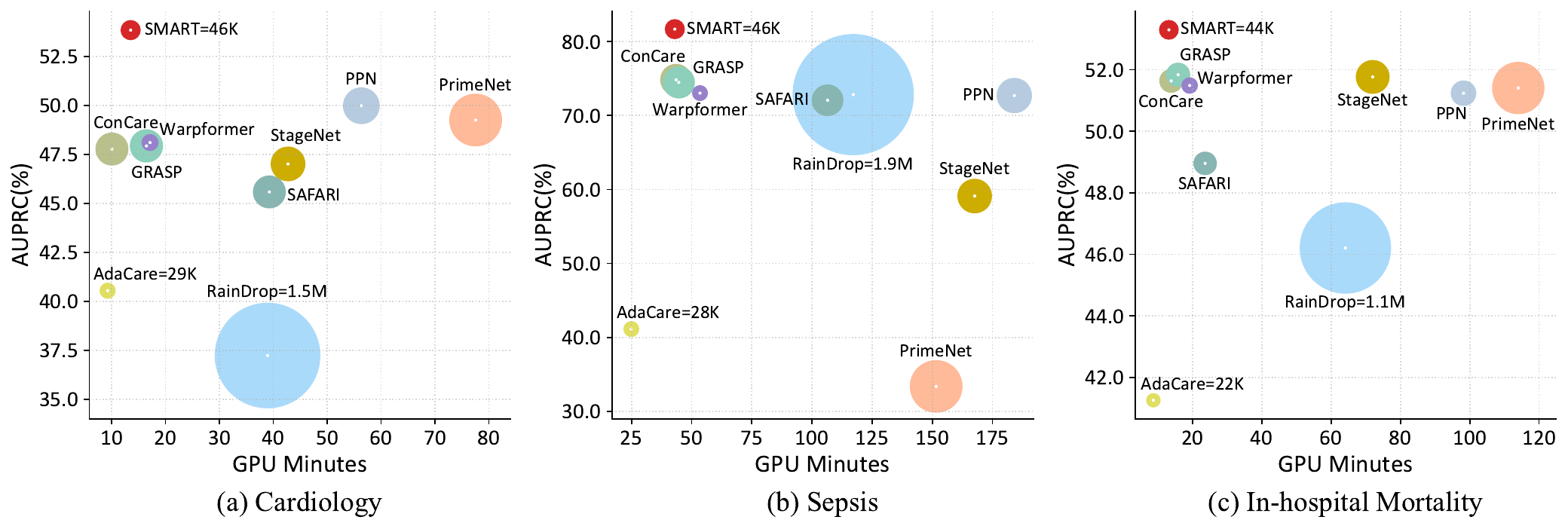}
    \caption{Training time, parameters, and AUPRC(\%) of all models on the three datasets. The size of the circle represents the number of parameters. The GPU runtime is counted at the start of training.}
    \label{efficiency}
\end{figure}
We conduct runtime and parameter comparisons between our method and baselines on the Cardiology, Sepsis, and in-hospital mortality prediction tasks. The number of parameters, the averaged runtime (minutes) on GPU (counted at the start of training), and the AUPRC of different methods are reported in Figure \ref{efficiency} for comparisons. Since the Sepsis dataset contains a larger number of patients, the model trained on it takes longer on average. As illustrated, RainDrop has the largest number of parameters and its training time is several times that of some of the models, such as AdaCare, ConCare, GRASP, Warpformer, and SMART, yet its prediction performance is not satisfactory and unstable on different datasets. StageNet, PPN, and PrimeNet achieve competitive performance on some of the datasets, however their training times are very long. Although AdaCare has the fewest parameters and the shortest training time, it also has one of the worst performances. In a nutshell, SMART is lightweight, fast, and reliable among all the models, reaching the best performance with only fewer parameters and shorter training time, which confirms its efficiency and effectiveness. Nevertheless, it is worth mentioning that due to the temporal attention mechanism in SMART performs in the visit dimension, its training time may grow quadratically as the number of visits increases.

\section{Conclusions}
In this work, we propose SMART, a self-supervised pre-trained model that is designed to handle the challenges of missingness and irregularity in EHR data. We introduce a two-stage training strategy that integrates imputing missingness in the representation space to enhance performance on clinical tasks. We elaborate on a novel MART block that captures temporal and variable interactions by introducing masks into the attention mechanism. Extensive experiments on six EHR tasks demonstrate that SMART outperforms existing baselines. The ablation study shows that all components in SMART contribute to the improvement. We further display that SMART is lightweight, efficient, and robust to missing values and achieves stable performance across different missing rates. In the future, we plan to provide more insights into the decision-making process and make it more explainable. Our code is available at \url{https://github.com/yzhHoward/SMART}.
{
\small
\bibliography{reference}
\bibliographystyle{unsrt}
}


\appendix

\section{Additional Experimental Results}
\subsection{Details of Datasets}
\label{sec:dataset}
When preprocessing all the datasets, we conduct z-score normalizations (subtract the mean and divide the deviation of the observed values) on the continuous features and one-hot encoding on the categorical features. We align the observation times to hours and pad the missing values with zeros. We use the first 48 hours of data for the Cardiology and the MIMIC-III mortality prediction dataset and the first 60 visits for the Sepsis dataset. We refer to the procedures from Harutyunyan et al. \cite{Harutyunyan2019} when generating samples from the MIMIC-III dataset. We split the data into training, validation, and test sets with a ratio of 8:1:1. The Cardiology dataset can be obtained at \url{https://physionet.org/content/challenge-2012/1.0.0/}. The Sepsis dataset can be obtained at \url{https://physionet.org/content/challenge-2019/1.0.0/}. The MIMIC-III dataset can be obtained at \url{https://physionet.org/content/mimiciii/1.4/}. Please follow the instructions on the PhysioNet website to access the data. We provide more details about the tasks on the MIMIC-III dataset as follows:

\textbf{In-hospital Mortality:} The goal of this task is to predict whether a patient will die at the end of hospitalization using the first 48 hours of data. We extract 21,139 records each containing 17 physiological variables with 43.3\% observed rate. There are 13.2\% of patients who have positive labels. In this task, the model can only observe the first 48 hours, which is advantageous from a practical point of view, as the earlier the clinician identifies the risk, the more timely the intervention can be implemented.

\textbf{Phenotyping:} Phenotyping can be applied in cohort construction for clinical studies, comorbidity detection and risk adjustment, quality improvement and surveillance, diagnosis and treatment planning. There are 25 different phenotypes in the dataset. We extract 41,903 patients on this task with 41.8\% observed rate.

\textbf{Decompensation:} The objective is to determine whether a patient will decease in the next 24 hours based on the data within a 24-hour time window. This can be utilized as an early warning in ICUs. We extract 41,744 patients The observed rate is 43.2\% and the positive rate is 3.5\%.

\textbf{Length of Stay:} The target of length-of-stay prediction is to predict the length of stay of a patient in the ICU. This task provides a more fine-grained view of the physiological states of the patients, assisting clinicians in monitoring the progress of the disease. We obtain 33,360 patients with 43.8\% observed rate. The length of stay is discretized into 10 bins, where bins 1-8 correspond to 1-8 days of stay, respectively, bin 8 for more than eight days but less than two weeks of stay, and bin 9 for living over two weeks.

\subsection{Implementation Details}
\label{sec:implementation}
All experiments of this model are carried out on a Linux server equipped with RTX 2080Ti GPUs. PyTorch 2.1.2 and CUDA 12.1 deep learning libraries are applied to build and train our neural network. While training models, we apply Adam optimizer \cite{kingma2014adam} with learning rate 1e-3. We train all the models (including baselines) with the same optimizer, learning rate, and a total batch size of 256. The training epochs and hyperparameters (such as dropout and hidden dimensions) of baselines are tuned for better performance. To evaluate precisely, we repeat every experiment 3 times with different random seeds (1, 42, and 3407). The standard deviation of the experimental results appears to be slightly larger because the division of the training, validation, and test sets is related to the seed rather than being fixed.

\textbf{Hyperparameters:} The hidden size $d$ of all modules in SMART is 32. The heads of the multi-head attention computation in temporal attention and variable attention are set to 4. The layer number $L$ of the MART blocks is 2. The model is pre-trained for 25 epochs and fine-tuned for 25 epochs. The unfreeze epoch is set to 5. We employ the same hyperparameter configurations across the datasets in our experiments. The probability interval $p$ for generating masks is set to (0, 0.75), and the dropout rate is set to 0.1. The exponential moving average (EMA) decay rate is set to 0.996.

\begin{table}
    \caption{Performance comparison of AUROC and min(Se, P+) with standard deviation on the four clinical tasks. The best results are marked in bold. The second best results are underlined.}
    \label{more_results}
    \centering
    \begin{tabular}{l|cccc}
        \toprule
        \multirow{2}{*}{\textbf{Model}} & \multicolumn{2}{c}{\textbf{Cardiology}} & \multicolumn{2}{c}{\textbf{Sepsis}} \\
        & AUROC & min(Se, P+) & AUROC & min(Se, P+) \\
        \midrule
        AdaCare & 78.01{\scriptsize $\pm$1.16} & 41.57{\scriptsize $\pm$3.60} & 84.70{\scriptsize $\pm$0.43} & 42.88{\scriptsize $\pm$2.47} \\
        StageNet & 82.55{\scriptsize $\pm$1.75} & \underline{49.91{\scriptsize $\pm$1.08}} & 91.25{\scriptsize $\pm$1.50} & 56.21{\scriptsize $\pm$3.08} \\
        ConCare & 82.74{\scriptsize $\pm$1.47} & 46.65{\scriptsize $\pm$2.71} & 94.39{\scriptsize $\pm$0.74} & 69.79{\scriptsize $\pm$3.07} \\
        GRASP & 82.98{\scriptsize $\pm$0.66} & 49.53{\scriptsize $\pm$3.23} & 94.43{\scriptsize $\pm$0.76} & 69.75{\scriptsize $\pm$2.10} \\
        SAFARI & 81.84{\scriptsize $\pm$1.09} & 47.54{\scriptsize $\pm$2.32} & \underline{94.64{\scriptsize $\pm$0.83}} & \underline{70.03{\scriptsize $\pm$1.68}} \\
        PPN & 83.42{\scriptsize $\pm$1.23} & 48.45{\scriptsize $\pm$1.51} & 94.25{\scriptsize $\pm$1.07} & 68.96{\scriptsize $\pm$2.25} \\
        RainDrop & 76.86{\scriptsize $\pm$1.67} & 38.31{\scriptsize $\pm$1.41} & 92.90{\scriptsize $\pm$0.79} & 68.74{\scriptsize $\pm$2.42} \\
        Warpformer & 83.21{\scriptsize $\pm$2.09} & 48.73{\scriptsize $\pm$3.22} & 93.86{\scriptsize $\pm$0.77} & 65.93{\scriptsize $\pm$1.79} \\
        PrimeNet & \underline{84.19{\scriptsize $\pm$0.52}} & 49.67{\scriptsize $\pm$2.58} & 83.60{\scriptsize $\pm$1.16} & 36.18{\scriptsize $\pm$2.39} \\
        \midrule
        SMART & \textbf{85.80{\scriptsize $\pm$0.05}} & \textbf{52.63{\scriptsize $\pm$1.37}} & \textbf{95.92{\scriptsize $\pm$0.84}} & \textbf{75.61{\scriptsize $\pm$2.76}} \\
        \midrule
        \multirow{2}{*}{\textbf{Model}} & \multicolumn{2}{c}{\textbf{In-hospital Mortality}} & \multicolumn{2}{c}{\textbf{Decompensation}}\\
        & AUROC & min(Se, P+) & AUROC & min(Se, P+)\\
        \midrule
        AdaCare & 78.35{\scriptsize $\pm$1.12} & 42.68{\scriptsize $\pm$0.81} & 87.06{\scriptsize $\pm$1.23} & 49.96{\scriptsize $\pm$1.95}\\
        StageNet & \underline{85.32{\scriptsize $\pm$0.62}} & 49.84{\scriptsize $\pm$3.12} & 92.49{\scriptsize $\pm$0.47} & 60.44{\scriptsize $\pm$0.93} \\
        ConCare & 84.97{\scriptsize $\pm$0.96} & 49.61{\scriptsize $\pm$1.61} & 92.43{\scriptsize $\pm$0.46} & 61.54{\scriptsize $\pm$1.01} \\
        GRASP & 84.79{\scriptsize $\pm$1.26} & 49.68{\scriptsize $\pm$2.19} & 92.25{\scriptsize $\pm$0.34} & 60.76{\scriptsize $\pm$1.54} \\
        SAFARI & 84.48{\scriptsize $\pm$1.16} & 48.81{\scriptsize $\pm$1.38} & 90.69{\scriptsize $\pm$0.16} & 58.64{\scriptsize $\pm$1.73} \\
        PPN & 84.90{\scriptsize $\pm$0.96} & 50.42{\scriptsize $\pm$1.12} & 91.50{\scriptsize $\pm$0.70} & 60.65{\scriptsize $\pm$0.91} \\
        RainDrop & 83.21{\scriptsize $\pm$0.99} & 46.30{\scriptsize $\pm$1.51} & 90.48{\scriptsize $\pm$1.60} & 56.55{\scriptsize $\pm$2.44} \\
        Warpformer & 84.11{\scriptsize $\pm$1.10} & 48.35{\scriptsize $\pm$2.02} & 93.24{\scriptsize $\pm$0.96} & 64.42{\scriptsize $\pm$0.97} \\
        PrimeNet & 84.82{\scriptsize $\pm$1.05} & \underline{50.96{\scriptsize $\pm$1.01}} & 91.15{\scriptsize $\pm$1.49} & 54.99{\scriptsize $\pm$2.66} \\
        \midrule
        SMART & \textbf{85.61{\scriptsize $\pm$0.62}} & \textbf{51.29{\scriptsize $\pm$1.35}} & \textbf{94.20{\scriptsize $\pm$0.12}} & \textbf{65.52{\scriptsize $\pm$1.74}} \\
        \bottomrule
    \end{tabular}
\end{table}

\subsection{Experimental Results on More Metrics}
\label{sec:more_results}
In addition to AUPRC and F1 Score, we also provide the results of other metrics on the binary classification task, such as the area under the receiver operating characteristic curve (AUROC), which is a widely used metric in the healthcare field. We also provide the results of the minimum of the precision and sensitivity (min(Se, P+)), which is often used in the healthcare field \cite{zhang2021grasp}. The results are shown in Table \ref{more_results}. The results illustrate that SMART outperforms existing baselines across all metrics on the four clinical tasks. AUROC can be used to examine the model's ability to distinguish between positive and negative samples, while min(Se, P+) can be used to evaluate the model's ability to balance sensitivity and precision. By observing the results, we can find that the results on these metrics are consistent with the results on AUPRC and F1 Score. Specifically, AdaCare performs the worst in the baseline, and StageNet, RainDrop, and PrimeNet show unstable performance on some datasets. In summary, SMART achieves the best performance on all metrics, demonstrating its effectiveness in learning representations and predicting patient health status.

\begin{table}
    \caption{Ablation study of mask information in SMART on the Cardiology, Sepsis, and MIMIC-III in-hospital mortality.}
    \label{mask_ablation}
    \centering
    \small
    \resizebox{\linewidth}{!}{
    \begin{tabular}{l|cccccc}
        \toprule
        \multirow{2}{*}{\textbf{SMART}} & \multicolumn{2}{c}{\textbf{Cardiology}} & \multicolumn{2}{c}{\textbf{Sepsis}} & \multicolumn{2}{c}{\textbf{In-hospital Mortality}}\\
        & AUPRC(\%) & F1 Score(\%) & AUPRC(\%) & F1 Score(\%) & AUPRC(\%) & F1 Score(\%)\\
        \midrule
        Full & \textbf{53.84{\scriptsize $\pm$2.24}} & \textbf{47.53{\scriptsize $\pm$2.33}} & \textbf{81.67{\scriptsize $\pm$0.84}} & \textbf{75.37{\scriptsize $\pm$2.62}} & \textbf{53.30{\scriptsize $\pm$0.12}} & \textbf{44.23{\scriptsize $\pm$2.03}} \\
        \midrule
        w/o Mask & 49.03{\scriptsize $\pm$1.93} & 41.98{\scriptsize $\pm$1.71} & 76.35{\scriptsize $\pm$2.76} & 69.51{\scriptsize $\pm$2.11} & 50.43{\scriptsize $\pm$2.44} & 39.70{\scriptsize $\pm$1.60} \\
        \midrule
        w/o Mask in Encoder & 49.74{\scriptsize $\pm$2.43} & 42.70{\scriptsize $\pm$2.11} & 77.37{\scriptsize $\pm$2.53} & 71.96{\scriptsize $\pm$2.26} & 52.29{\scriptsize $\pm$1.06} & 42.65{\scriptsize $\pm$2.91} \\
        w/o Mask in Temporal Attention & 53.71{\scriptsize $\pm$2.22} & 46.69{\scriptsize $\pm$2.35} & 81.08{\scriptsize $\pm$2.97} & 75.62{\scriptsize $\pm$2.83} & 53.29{\scriptsize $\pm$1.58} & 42.83{\scriptsize $\pm$3.09} \\
        w/o Mask in Variable Attention & 52.37{\scriptsize $\pm$3.17} & 44.53{\scriptsize $\pm$1.83} & 76.87{\scriptsize $\pm$2.20} & 72.06{\scriptsize $\pm$2.66} & 52.97{\scriptsize $\pm$0.36} & 41.21{\scriptsize $\pm$2.99} \\
        \bottomrule
    \end{tabular}
    }
\end{table}

\subsection{Ablation Study on the Mask Information}
We conduct an ablation study on the mask information in SMART. This ablation study can be regard as an extension of the reduced version \textbf{w/o Mask} in Table \ref{ablation} and it is a deeper exploration of how the mask information affects the performance of SMART. Specifically, we remove the mask information in the input encoder and the attentions in the MART blocks separately, namely \textbf{w/o Mask in Encoder}, \textbf{w/o Mask in Temporal Attention}, and \textbf{w/o Mask in Variable Attention}. The results are shown in Table \ref{mask_ablation}. We observe that the performance of SMART drops significantly when the mask information is removed in the input encoder, which indicates that the mask information in the input encoder is critical to learn complete representations. When the mask information is removed in the attentions, the performance of SMART also decreases, which demonstrates that the mask information in the attentions is essential to resist noise when capturing the temporal and variable interactions. These results of the ablation study on the mask information further confirm each design in SMART is indispensable and contributes to the improvement.

\begin{table}
    \caption{Study on the layer numbers of the MART block on the Cardiology, Sepsis, and in-hospital mortality. The best results are marked in bold.}
    \label{layers}
    \centering
    \small
    \resizebox{\linewidth}{!}{
    \begin{tabular}{c|cccccc}
        \toprule
        \textbf{SMART} & \multicolumn{2}{c}{\textbf{Cardiology}} & \multicolumn{2}{c}{\textbf{Sepsis}} & \multicolumn{2}{c}{\textbf{In-hospital Mortality}}\\
        $L$ & AUPRC(\%) & F1 Score(\%) & AUPRC(\%) & F1 Score(\%) & AUPRC(\%) & F1 Score(\%)\\
        \midrule
        1 & 52.66{\scriptsize $\pm$2.67} & 44.07{\scriptsize $\pm$2.72} & 81.26{\scriptsize $\pm$2.68} & 74.68{\scriptsize $\pm$2.92} & 53.40{\scriptsize $\pm$1.66} & 42.55{\scriptsize $\pm$2.83} \\
        2 & \textbf{53.84{\scriptsize $\pm$2.24}} & \textbf{47.53{\scriptsize $\pm$2.33}} & \textbf{81.67{\scriptsize $\pm$0.84}} & \textbf{75.37{\scriptsize $\pm$2.62}} & 53.30{\scriptsize $\pm$0.12} & 44.23{\scriptsize $\pm$2.03} \\
        3 & 52.18{\scriptsize $\pm$1.42} & 46.54{\scriptsize $\pm$1.88} & 81.35{\scriptsize $\pm$3.24} & 74.30{\scriptsize $\pm$2.74} & \textbf{53.82{\scriptsize $\pm$1.51}} & 43.28{\scriptsize $\pm$3.53} \\
        4 & 50.56{\scriptsize $\pm$3.90} & 46.65{\scriptsize $\pm$1.75} & 79.98{\scriptsize $\pm$2.91} & 73.68{\scriptsize $\pm$2.02} & 51.89{\scriptsize $\pm$0.71} & \textbf{44.97{\scriptsize $\pm$2.62}} \\
        \bottomrule
    \end{tabular}
    }
\end{table}

\begin{table}
    \caption{Study on the mask ratios of SMART in the pre-training stage on the Cardiology, Sepsis, and in-hospital mortality. The best results are marked in bold.}
    \label{mask}
    \centering
    \small
    \resizebox{\linewidth}{!}{
    \begin{tabular}{c|cccccc}
        \toprule
        \textbf{SMART} & \multicolumn{2}{c}{\textbf{Cardiology}} & \multicolumn{2}{c}{\textbf{Sepsis}} & \multicolumn{2}{c}{\textbf{In-hospital Mortality}}\\
        $p$ & AUPRC(\%) & F1 Score(\%) & AUPRC(\%) & F1 Score(\%) & AUPRC(\%) & F1 Score(\%)\\
        \midrule
        (0, 0.25) & 52.81{\scriptsize $\pm$2.63} & 44.05{\scriptsize $\pm$2.19} & 80.78{\scriptsize $\pm$2.46} & 74.46{\scriptsize $\pm$2.95} & 52.55{\scriptsize $\pm$0.39} & 44.17{\scriptsize $\pm$3.91} \\
        (0, 0.5) & 53.77{\scriptsize $\pm$2.45} & 46.92{\scriptsize $\pm$0.61} & 80.89{\scriptsize $\pm$2.23} & 74.04{\scriptsize $\pm$2.75} & 53.41{\scriptsize $\pm$0.78} & \textbf{44.90{\scriptsize $\pm$3.56}} \\
        (0, 0.75) & \textbf{53.84{\scriptsize $\pm$2.24}} & \textbf{47.53{\scriptsize $\pm$2.33}} & \textbf{81.67{\scriptsize $\pm$0.84}} & \textbf{75.37{\scriptsize $\pm$2.62}} & 53.30{\scriptsize $\pm$0.12} & 44.23{\scriptsize $\pm$2.03} \\
        (0.25, 0.75) & 53.66{\scriptsize $\pm$2.64} & 45.06{\scriptsize $\pm$3.09} & 81.82{\scriptsize $\pm$2.04} & 74.71{\scriptsize $\pm$3.03} & \textbf{53.48{\scriptsize $\pm$0.66}} & 43.03{\scriptsize $\pm$3.00} \\
        (0.5, 0.75) & 53.63{\scriptsize $\pm$2.38} & 46.00{\scriptsize $\pm$1.64} & 81.69{\scriptsize $\pm$2.02} & 74.87{\scriptsize $\pm$2.29} & 53.37{\scriptsize $\pm$1.00} & 42.45{\scriptsize $\pm$2.39} \\
        \bottomrule
    \end{tabular}
    }
\end{table}

\begin{table}
    \caption{Performance comparison with standard deviation on Cardiology, Sepsis, and in-hospital mortality with basic baselines.}
    \label{basic}
    \centering
    \small
    \resizebox{\linewidth}{!}{
    \begin{tabular}{l|cccccc}
        \toprule
        \multirow{2}{*}{\textbf{Model}} & \multicolumn{2}{c}{\textbf{Cardiology}} & \multicolumn{2}{c}{\textbf{Sepsis}} & \multicolumn{2}{c}{\textbf{In-hospital Mortality}}\\
        & AUPRC(\%) & F1 Score(\%) & AUPRC(\%) & F1 Score(\%) & AUPRC(\%) & F1 Score(\%)\\
        \midrule
        GRU & 47.10{\scriptsize $\pm$3.56} & 37.84{\scriptsize $\pm$0.92} & 49.73{\scriptsize $\pm$3.12} & 47.66{\scriptsize $\pm$2.15} & 50.68{\scriptsize $\pm$2.07} & 41.70{\scriptsize $\pm$2.21} \\
        Transformer & 42.92{\scriptsize $\pm$3.70} & 42.82{\scriptsize $\pm$3.73} & 22.56{\scriptsize $\pm$4.41} & 24.75{\scriptsize $\pm$3.90} & 43.07{\scriptsize $\pm$3.68} & 42.94{\scriptsize $\pm$3.36} \\
        \midrule
        SMART & \textbf{53.84{\scriptsize $\pm$2.24}} & \textbf{47.53{\scriptsize $\pm$2.33}} & \textbf{81.67{\scriptsize $\pm$0.84}} & \textbf{75.37{\scriptsize $\pm$2.62}} & \textbf{53.30{\scriptsize $\pm$0.12}} & \textbf{44.23{\scriptsize $\pm$2.03}} \\
        \bottomrule
    \end{tabular}
    }
\end{table}

\subsection{Study on the Hyperparameters}
To explore the affect of the hyperparameters on the performance of SMART, we conduct a study on the layer numbers of the MART block and the mask ratios in the pre-training stage. The results are shown in Table \ref{layers} and Table \ref{mask}. On the layer numbers of the MART block, the results show that the performance of SMART is the best when the layer numbers are set to 2. Besides, we find that too much layer numbers may lead to decreased performance. 

When it comes to the mask ratios in the pre-training stage, when the mask ratios are set to (0, 0.75), SMART achieves the best performance on the Cardiology and Sepsis, whereas the mask ratio $p=$(0.25, 0.75) obtains the best AUPRC and $p=$(0, 0.5) achieves the best F1 Score on the in-hospital mortality.
These experiments shed light on the importance of hyperparameters in the performance of SMART, and provide guidance for the selection of hyperparameters in practice.
\label{sec:hyperparameters}

\subsection{Comparisons with Basic Baselines}
We compare SMART with basic baselines on the Cardiology, Sepsis, and in-hospital mortality tasks, including gated recurrent units (GRU) and Transformer \cite{vaswani2017attention}. The results are shown in Table \ref{basic}. We observe that SMART outperforms the basic baselines on all datasets, which shows the effectiveness of SMART in learning representations and predicting patient health status. The results also show that the basic baselines have poor performance especially on the Sepsis datasets, which may be due to the complexity and higher missing rate of the dataset.

\section{Limitations}
\label{sec:limitations}
This work focuses on multivariate time-series EHR data in ICU scenarios, mainly including patients' physiologic indications and laboratory tests. In fact, patients' EHR data may be more diversified, such as demographics, diagnosis (ICD codes), surgery (ICD codes), medication (ATC codes), radiographs, ECGs, and clinical notes (e.g., examination reports, nursing notes, etc.), and the above-mentioned data are beyond the scope of this work. In addition, the studies in this work do not address interpretability, which may make it more difficult for physicians to understand the predictions. Besides, we evaluated the experimental results using only three large datasets in reference to previous work and did not separately evaluate the results on smaller datasets. Smaller datasets may lead to poorer pre-training results as the model is not able to fully learn the feature associations of the data. Like all models with temporal attention, such as ConCare \cite{ma2020concare} and Warpformer \cite{zhang2023warpformer}, the time and space complexity of our model is $O(T^2)$, which means that the model may face slow speed or insufficient memory when the input sequence is too long.

\section{Broader Impact}
\label{sec:impact}
The proposed model is designed to improve the representation learning capability of EHR data in ICU scenarios. The model can be used to predict patients' health status, such as in-hospital mortality, sepsis, and other diseases. The model can be used to assist physicians in making clinical decisions, such as early warning of patients' health status, which may help reduce the mortality rate of patients. The model can also be used to assist in the allocation of medical resources, such as the allocation of ICU beds, which may help reduce the burden on the healthcare system. However, the model may also have some negative impacts. For example, the model's predictions may be biased, which may result in patients being treated unfairly. Additionally, there may be ethical issues with this model, as it may lead to decisions that are not in the best interest of the patient. Therefore, it is important to carefully evaluate the model before actually using it. It should be noted that the model cannot replace doctors in making decisions. The model is only a tool to assist doctors in making decisions. The model should be used in conjunction with doctors to make the best decisions for patients.

\end{document}